\documentclass[runningheads]{llncs}

\usepackage{eccv}

\usepackage{url}
\usepackage{multirow}

\usepackage{eccvabbrv}

\usepackage{graphicx}
\usepackage{booktabs}
\usepackage{algorithm}
\usepackage{algorithmic}
\usepackage{amsmath}
\usepackage[table,dvipsnames]{xcolor}
\usepackage[numbers]{natbib}
\usepackage[accsupp]{axessibility}

\usepackage{pifont}

\definecolor{darkblue}{RGB}{0,0,160}
\definecolor{royalblue}{RGB}{0,90,255}

\usepackage[pagebackref,breaklinks,colorlinks,citecolor=eccvblue]{hyperref}

\usepackage{orcidlink}

\begin{document}

\title{PhasorNet: Learning Structure from
Frequency for Real-Time Stereo Matching} 

\titlerunning{PhasorNet}

\author{Md Raqib Khan\inst{1} \and
Santosh Kumar Vipparthi\inst{2} \and
Subrahmanyam Murala\inst{1}}

\authorrunning{Khan et al.}


\institute{CVPR Lab, Trinity College Dublin, The University of Dublin, Dublin, Ireland
\and
CVPR Lab, Indian Institute of Technology Ropar, Rupnagar, Punjab, India
\email{khanmd@tcd.ie}
}

\maketitle

\begin{abstract}
Accurate stereo matching remains challenging in ill-posed regions such as fine structures, reflective, or transparent objects, where appearance cues are often ambiguous or unreliable. To tackle this, we propose PhasorNet, a lightweight yet powerful framework that boosts geometric discrimination via frequency-domain cues. At its core, the Phase-Augmented Transformer (PAT) injects Fourier-derived phase information into the attention mechanism, yielding photometrically robust, structure-preserving features
that prioritize structural consistency in difficult areas. Additionally, we develop a Geometry-Context Fusion Refinement Module (GCFRM) that combines a full-resolution convolutional stream with a lightweight attention-based stream (leveraging WQA and CDGA blocks) to efficiently preserve fine details and object boundaries without excessive overhead. Training is further enhanced by a multi-scale Edge-guided High-Error Region (EHR) loss that adaptively focuses optimization on high-error and edge regions, guiding hierarchical cost volume refinement. With only 5.3M parameters, PhasorNet achieves state-of-the-art performance on the challenging ETH3D benchmark while exhibiting excellent cross-domain generalization on KITTI, delivering an efficient and practical solution for accurate real-time stereo matching.
\end{abstract}

\begin{figure}[!htb]
    \centering
    \includegraphics[width=\linewidth]{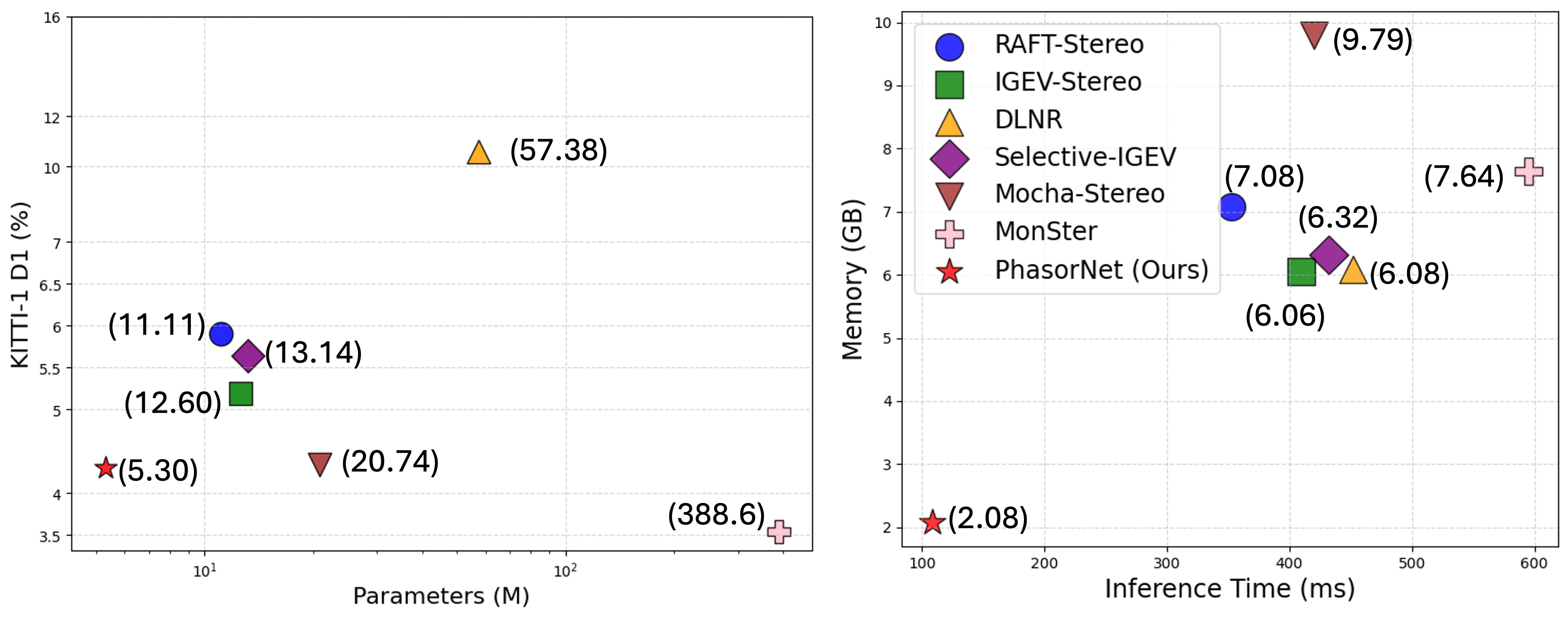}
    \caption{Accuracy-efficiency trade-offs of state-of-the-art stereo matching methods trained on the Scene Flow dataset. Left: D1 error versus model size (number of parameters). Right: Memory consumption versus inference time. PhasorNet is highlighted, illustrating a favorable balance between accuracy and computational efficiency.}
    \label{fig:motivation}
\end{figure}

\begin{figure*}[!htb]
    \centering
    \includegraphics[width=1\linewidth]{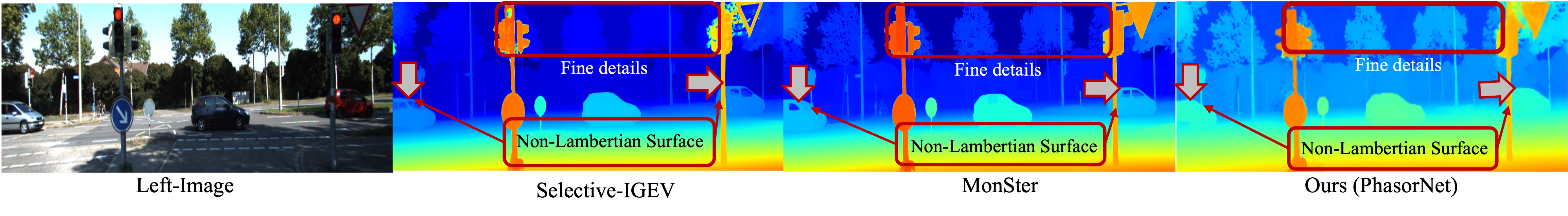}
    \captionsetup{skip=3pt}

    \caption{Example Illustrating challenges in stereo matching. Non-Lambertian and reflective regions (e.g., vehicle glass) and fine structural details lead to photometric inconsistencies and weak texture, causing recent methods such as Selective-IGEV and MonSter to produce degraded or smoothed disparities. This motivates the need for structure-aware feature representations that emphasize geometric consistency over appearance similarity.}
    \label{fig:motivation1}
\end{figure*}

\section{Introduction}

Accurate stereo depth estimation is fundamental for 3D perception in autonomous driving, robotics, and augmented reality. This task requires establishing precise pixel‑wise correspondences between rectified image pairs to recover scene geometry. Despite significant advances, contemporary learning‑based methods face three interconnected challenges: learning geometrically informative features, efficiently processing matching costs, and refining disparities while preserving fine details under practical computational constraints. The foundation of any stereo system lies in its feature representation. While deep learning has largely replaced handcrafted features with learned representations, conventional backbones whether CNN‑based (ResNet~\cite{he2016deep}), Transformer‑based (ViT~\cite{dosovitskiy2020image}) are typically optimized for semantic recognition rather than geometric matching. Consequently, extracted features may be semantically rich but geometrically ambiguous, especially on challenging surfaces such as transparent glass, textureless walls, or reflective materials~\cite{zhao2023high,wang2022antioversmooth}. As illustrated in Fig.~\ref{fig:motivation1}, transparent surfaces severely disrupt appearance‑based matching, motivating the need for features that capture invariant structural contours.

To address this, we introduce the \textit{Phase‑Augmented Transformer (PAT)}, whose core is a \textit{Phase‑Augmented Cross‑Attention (PAC)} module (defined in Sec.~\ref{sec:pat}). Unlike intensity‑based features, which are sensitive to transparency, reflections, and low texture, PAT leverages phase‑aware attention to capture high‑frequency structural cues in the Fourier domain. The phase of the Fourier transform, when isolated via magnitude normalisation, is largely insensitive to intensity variations and preserves the dominant geometric structure of the signal~\cite{oppenheim1981importance}. Furthermore, the relationship between phases across views directly encodes spatial displacements, a property widely exploited in frequency‑domain registration~\cite{kuglin1975phase,foroosh2002extension}. This makes phase information inherently suited for robust matching in ill‑posed regions. For feature extraction, the global nature of the Fourier transform is ideal: it captures large‑scale geometric consistency (e.g., building edges, car silhouettes) in a manner robust to local intensity variations.

Once features are extracted, conventional approaches construct a 3D cost volume processed by expensive 3D convolutions~\cite{chang2018pyramid,shen2021cfnet}, incurring high memory and computation. Recent methods reduce cost via adaptive 2D aggregation~\cite{xu2020aanet} or iterative refinement~\cite{lipson2021raft,li2022practical}, but often sacrifice edge precision or speed. Departing from this paradigm, we eliminate 3D convolutions entirely and process the cost volume at multiple scales using only 2D operations. To guide learning, we propose a novel \textit{Edge‑guided and High‑Error Region (EHR) Loss} that supervises predicted disparity maps at multiple scales, emphasizing object boundaries and hard regions without heavy post‑processing.

The final stage refines the initial disparity estimate. While iterative methods like RAFT ‑ Stereo \cite{lipson2021raft} achieve high accuracy, they rely on multiple recurrent updates, increasing latency and often over‑smoothing object boundaries. To address this, we introduce a \textit{Geometry‑Context Fusion Refinement Module (GCFRM)} that combines a full‑resolution convolutional stream (preserving spatial details) with a lightweight attention‑based stream that captures global context. The attention stream integrates \textit{Wavelet‑Based Query Attention (WQA)} and \textit{Cross ‑ Dimensional Gated Attention (CDGA)}. Unlike the Fourier transform used in PAT for global phase cues, the discrete wavelet transform (DWT) in WQA provides \textit{localised} multi‑scale frequency decomposition, which is essential for sharp boundary preservation during refinement. Thus, the two transforms are deliberately chosen for complementary
roles: FFT for global structural cues robust to photometric variations,
and DWT for local edge enhancement, this design is validated by ablations.

Our main contributions are:
\begin{itemize}
    \item \textit{PhasorNet}, a compact end‑to‑end stereo network (5.3M parameters) that balances accuracy, efficiency, and detail preservation.
    \item A Phase-Augmented Transformer encoder that injects Fourier phase information into attention, significantly improving robustness on transparent, reflective, and low-texture surfaces.
   \item A Geometry-Context Fusion Refinement Module that combines a full-resolution convolutional stream with lightweight attention (WQA and CDGA) to preserve fine details and boundaries.
    \item The EHR Loss, a multi‑scale supervision strategy that emphasizes edges and high‑error regions to guide hierarchical cost volume refinement.
\end{itemize}

Extensive experiments on Scene Flow, KITTI 2012/2015, and ETH3D demonstrate that PhasorNet achieves a compelling accuracy‑efficiency trade‑off, making it well‑suited for real‑world, low‑resource deployment. 

\section{Related Work}
\label{sec:RelatedWork}

Learning-based stereo matching progressed from global optimization~\cite{klaus2006segment,felzenszwalb2006efficient} to deep cost-volume methods. GC-Net~\cite{kendall2017end} and PSMNet~\cite{chang2018pyramid} popularized 3D convolutions for aggregation and regression. Later works improved efficiency (AANet~\cite{xu2020aanet} with adaptive 2D aggregation; CFNet~\cite{shen2021cfnet} with cascaded refinement) but many still incur high memory/compute costs from volumetric or multi-stage designs.

Robust correspondence features remain challenging. pretrained backbones like (ResNet~\cite{he2016deep}, EfficientNet~\cite{tan2019efficientnet}) often produce oversmoothed or semantically misaligned features~\cite{chen2024mocha}. Recent methods add geometric cues via attention~\cite{xu2022attention} or monocular priors. Vision foundation models enable stronger zero-shot generalization~\cite{yang2024depth}, yet cost volumes frequently require heavy post-processing. Transformer encoders improve context but risk losing high-frequency details critical at boundaries and textureless areas~\cite{wang2022antioversmooth}.

More recently foundation-model approaches leverage monocular priors from Depth Anything V2  for efficient, robust stereo matching. MonSter~\cite{cheng2025monster} integrates monodepth and stereo in a dual-branch architecture with confidence-guided mutual refinement, delivering excellent zero-shot generalization and superior handling of ill-posed regions with favorable computational trade-offs. DEFOM-Stereo~\cite{jiang2025defom} incorporates a robust monocular relative depth foundation model into a recurrent framework with combined encoders and scale updates, achieving top benchmark results but with higher memory/compute demands from recurrent updates and dense feature fusion.

Refinement architectures are crucial for recovering accurate, detail-preserving disparities. RAFT-Stereo \cite{lipson2021raft} and CREStereo \cite{li2022practical} leverage recurrent update operators across multiple iterations, achieving high accuracy but often at the cost of smoothed edges and increased inference time. To mitigate over-smoothing, DLNR \cite{zhao2023high} employs decoupled LSTM streams and disparity normalization.
\begin{figure*}[!htb]
    \centering
    \includegraphics[width=0.95\linewidth]{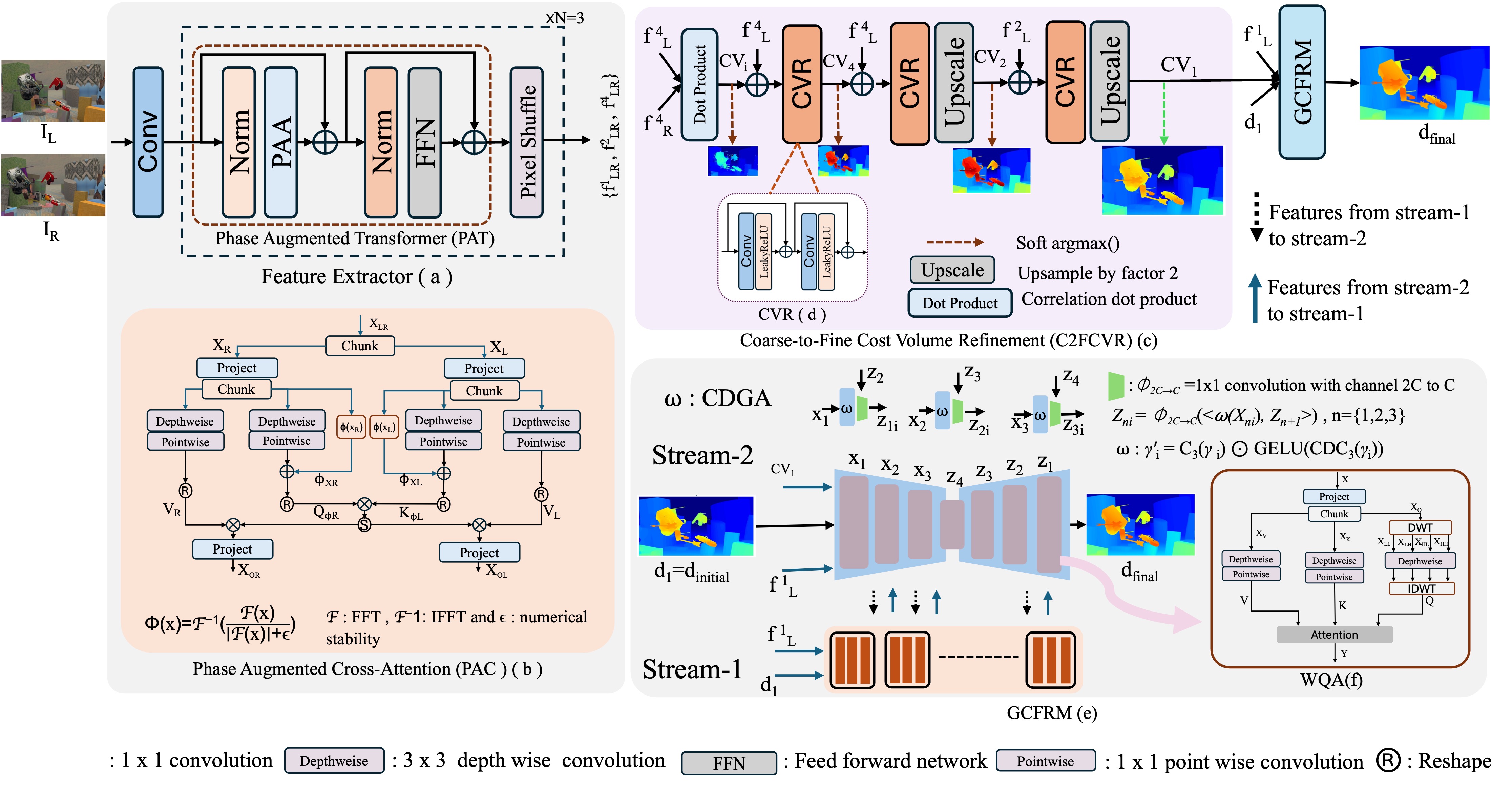}  
    \caption{Overview of the proposed PhasorNet architecture. 
    Given left and right input images $I_L$ and $I_R$, 
    (a) a phase-augmented transformer extracts multi-scale features from both views. 
    (b) An initial cost volume is constructed at coarse resolution and progressively refined in a coarse-to-fine manner using the proposed Coarse-to-Fine Cost Volume Refinement module (C2FCVR). 
    The resulting coarse disparity is further enhanced by the Geometry-Context Fusion Refinement Module (GCFRM (e))}
    \label{fig: PhasorNet}
\end{figure*}

\section{Methodology}
\label{sec:methodology}
\vspace{-1mm}
This section outlines our proposed architecture pipeline, as depicted in Fig.~\ref{fig: PhasorNet}. We detail the Phase-Augmented Transformer for feature extraction, Geometry-Context Fusion Refinement Module, and the multi-scale EHR Loss (Edge-guided and High-Error Region Loss) to refine the cost volume.

\subsection{Overall Pipeline}
\vspace{-1mm}
Given rectified stereo images \( I_L, I_R \in \mathbb{R}^{H \times W \times 3} \), our goal is to estimate a disparity map \( d_{final} \in \mathbb{R}^{H \times W} \) for per-pixel horizontal shifts. Our method comprises three components:

\textbf{Feature Extraction:} A Phase-Augmented Transformer Block generates multi-scale features \( f^i_{L}, f^i_{R} \in \mathbb{R}^{H/2^{i-1} \times W/2^{i-1} \times 2^{i-1}C} \) from both views.

\textbf{Cost Volume Construction and Refinement:}  
We construct an initial cost volume \( CV_i \) at \( \frac{1}{4} \) resolution and progressively refine it to \( \frac{1}{2} \) and full resolution (\( CV_2, CV_1 \)) using the Coarse-to-Fine Cost Volume Refinement (C2FCVR) module (Fig.~\ref{fig: PhasorNet}(c)). The matching cost is computed via a normalized inner product:
\begin{equation}
CV_{i}(d,h,w) = \frac{1}{N} \langle f_L^4(h,w),\, f_R^4(h,w-d) \rangle,
\end{equation}
where \( \langle \cdot,\cdot \rangle \) denotes the inner product and \( N \) is the number of channels.  

At each refinement stage, the cost volume is upsampled and fused with left-view features, followed by lightweight convolutional refinement with LeakyReLU activations (CVR module, Fig.~\ref{fig: PhasorNet}(d)) to enforce geometric consistency. This coarse-to-fine 2D refinement strategy, together with multi-scale EHR Loss supervision, effectively replaces computationally expensive 3D convolution-based cost aggregation while maintaining accurate disparity estimation, particularly in occluded and textureless regions.

\textbf{Disparity Refinement Module:} A Geometry-Context Fusion Refinement Module, incorporating Wavelet-Based Query Attention (WQA) and Cross Dimensional Gated Attention (CDGA), refines the initial disparity \( d_{init} =d_{1}\) into a final map \( d_{final} \in \mathbb{R}^{H \times W} \), preserving details in challenging areas with low computational cost.

\subsection{Phase-Augmented Cross-Attention (PAC)}
\label{sec:pat}

The Phase-Augmented Cross-Attention (PAC) module injects phase-consistent frequency-domain structural cues into the attention mechanism to enhance geometric correspondence estimation. Given left and right feature maps $ (X_L, X_R \in \mathbb{R}^{C \times H \times W}) $, phase-only reconstructions~\cite{oppenheim1981importance} are computed using magnitude normalized Fourier transforms:
\begin{equation}
\phi(x) =
\mathcal{F}^{-1}
\left(
\frac{\mathcal{F}(x)}
{\lvert \mathcal{F}(x) \rvert + \epsilon}
\right),
\end{equation}
where $(\mathcal{F}(\cdot))$ and $(\mathcal{F}^{-1}(\cdot))$ denote the 2D Fourier and inverse Fourier transforms, respectively, and $(\epsilon = 10^{-5})$ ensures numerical stability. By discarding magnitude information, $(\phi(\cdot))$ preserves high-frequency structural cues such as edges and contours while suppressing appearance-dependent variations~\cite{oppenheim1981importance}.

The reconstructed phase features are injected into the cross-view attention embeddings as residual augmentations:
\begin{equation}
Q_L' = Q_L + \phi(X_L), \qquad
K_R' = K_R + \phi(X_R),
\end{equation}
where $Q_L$ and $K_R$ denote the original query and key embeddings derived from the left and right views, respectively. The phase features are projected to the same embedding dimension before residual fusion. Phase augmentation is applied only to the query and key embeddings because they govern correspondence matching, whereas preserving unmodified value features retains richer contextual information during feature aggregation.

Unlike appearance features that vary under illumination or reflective changes, phase-only representations preserve structural alignment cues that remain consistent across stereo views, making them particularly effective for correspondence estimation in reflective and weakly textured regions. By augmenting attention with phase-consistent structural information, PAC improves geometric feature matching while remaining lightweight and computationally efficient.

Queries \((Q_L, Q_R)\), keys \((K_L, K_R)\), and values \((V_L, V_R)\) are generated from the input features \((X_L, X_R)\) using a lightweight projection composed of a \(1 \times 1\) pointwise convolution followed by a \(3 \times 3\) depthwise convolution. The cross-attention weights are computed as:
\begin{equation}
A = \mathrm{softmax} \left( \frac{Q_L' {K_R'}^\top}{\sqrt{d_k}} \cdot T \right),
\end{equation}
where \(d_k\) is the key dimensionality and \(T\) is a learnable temperature parameter.

The attention outputs are combined with the original features via residual connections:
\begin{align}
X_{OL} &= \psi_1(A \cdot V_L) + X_L, \\
X_{OR} &= \psi_1(A \cdot V_R) + X_R,
\end{align}
where \(\psi_1\) denotes a \(1 \times 1\) convolution. The resulting features are further refined using a lightweight feed-forward network with residual connections, following Restormer~\cite{Zamir2021Restormer}.

\subsection{Geometry-Context Fusion Refinement Module}
\label{sec:dualstream}
\vspace{-1mm}
\textit{GCFRM: Geometry-Context Fusion Disparity Refinement Module} enhances initial disparity predictions by integrating geometric and contextual cues. It takes as input the cost volume ($CV_1$), left image features ($f^1_{L}$), and the initial disparity map ($d_{init}$).
The module consists of two complementary streams:
    
\textbf{Convolutional Stream (Stream-1):}  
    This stream preserves fine details by processing the initial features $f_{L_1}$ and disparity $d_{init}$ through six stages. Each stage consists of three convolutional layers with ReLU activations (denoted as $\pi$). At each stage, its output is fused into the second stream via convolutional downsampling. Additionally, the stream receives fused features from the previous stage and upsampled features from the corresponding stage of the second stream (see GCFRM in Fig.~\ref{fig: PhasorNet}).
    \begin{equation}
    \begin{aligned}
    x_0 &= \pi\Bigl(\psi_1(\langle f^1_{L}, d_{init} \rangle)\Bigr), \\
    x_i &= \pi(\langle x_{i-1}, y_i \rangle), \quad i = 1, \dots, 6
    \end{aligned}
    \end{equation}

    Here, $x_0$ denotes the initial feature map of the convolutional stream, obtained by concatenating the left image features $f_{L_1}$ and the initial disparity $d_{init}$, passing them through a $1 \times 1$ convolution $\psi_1$, and applying three convolutional layers with ReLU activations (denoted by $\pi$). The operator $\langle \cdot, \cdot \rangle$ indicates channel-wise concatenation, and $x_i$ and $y_i$ represent the features of the first and second streams at stage $i$, respectively.

\textbf{Transformer-Based Encoder–Decoder Stream (Stream-2):}  
This stream generates the residual refinement by processing the left-view feature \(f^{1}_{L}\), the initial disparity \(d_{init}=d_1\), and the cost volume \(CV_1\) (see Fig.~\ref{fig: PhasorNet} (e)). It refines features through multi‑resolution attention while preserving local details and mitigating the effects of corrupted regions.  
Stream‑2 is composed of three encoder stages and three decoder stages, yielding six transformer blocks in total. Each of the six stages fuses one of the six multi‑scale feature maps produced by the convolutional Stream‑1 (\(x_0, \dots, x_5\)), creating a tight coupling between the two streams.

\noindent\textbf{Wavelet-Based Query Attention (WQA):}  
In WQA, queries are constructed from wavelet decomposed subbands of \(X_Q\) via a forward discrete wavelet transform (DWT) using a (db3) wavelet with zero‑padding \cite{daubechies1992ten}:
\begin{equation}
X_{LL}, X_{LH}, X_{HL}, X_{HH} = \text{DWT}(X_Q),
\end{equation}
which are then processed with depthwise separable convolutions (\(\psi\) with \(3\times3\) kernel):
\begin{equation}
\begin{split}
X_{LL}' = \psi_a(X_{LL}), \quad X_{LH}' = \psi_h(X_{LH}), \\
X_{HL}' = \psi_v(X_{HL}), \quad X_{HH}' = \psi_d(X_{HH}).
\end{split}
\end{equation}
The processed subbands are reassembled via the inverse DWT to form the query \(Q\). Applying wavelet decomposition exclusively to the query denoises and enhances structural cues~\cite{donoho1994threshold} that govern attention selection, while preserving the original contextual information in the keys and values. This asymmetric design improves robustness to noisy or corrupted regions without over‑filtering feature representations, and achieves a better accuracy‑efficiency trade‑off than a symmetric WQA (Table~\ref{tab:wqa_symmetric}).

Keys (\(K\)) and values (\(V\)) are derived from \(X_{K,V}\) through a \(1\times1\) convolution followed by a \(3\times3\) depthwise convolution, efficiently preserving fine details. The resulting attention is computed as:
\begin{equation}
\text{Y} = \text{softmax}\left( \frac{Q K^\top}{\sqrt{d}} \cdot T \right) V,
\end{equation}
where \(d\) is the feature dimension of the queries/keys and \(T\) is a learnable scaling factor~\cite{vaswani2017attention}. This formulation reduces the impact of noisy or corrupted regions on disparity refinement.

\noindent\textbf{Cross-Dimensional Gated Attention (CDGA):}  
CDGA propagates encoder features to the decoder through a cross‑dimensional \(3\times3\) convolution (CDC3), enabling selective feature fusion:
\begin{equation}
Z_i = C_3(a_i) \odot G(\mathrm{CDC3}(a_i)),
\end{equation}
where CDC3 applies dynamic attention across spatial, channel, filter, and kernel dimensions:
\begin{equation}
\gamma'_i = \sum_{i=1}^{n} \alpha_{w_i} \odot \alpha_{f_i} \odot \alpha_{c_i} \odot \alpha_{s_i} \odot W_i * \gamma_i.
\end{equation}
Here, \(\alpha_{w_i}\), \(\alpha_{f_i}\), \(\alpha_{c_i}\), and \(\alpha_{s_i}\) denote attention weights over convolution kernels, output filters, input channels, and spatial dimensions (\(k \times k\)), respectively~\cite{li2022omni}. By emphasizing informative features while suppressing unreliable responses, CDGA improves encoder–decoder fusion and enhances residual disparity refinement, particularly in occluded, reflective, and textureless regions, with minimal computational overhead.

\noindent\textbf{Overall Flow of Stream-2:}  
Stream-2 operates as a wavelet‑based encoder decoder that fuses features from Stream‑1. The encoder reduces spatial resolution over three stages, while the decoder symmetrically restores it.

\noindent\textbf{Input projection:}  
The cost volume \(CV_1\), left image features \(f^{1}_{L}\), and initial disparity \(d_{init}\) are concatenated and passed through a \(1\times1\) convolution, yielding the encoder input \(X_0\) at full resolution:
\[
X_0 = \psi_1\bigl( \langle CV_1,\, f^{1}_{L},\, d_{init} \rangle \bigr).
\]

\noindent\textbf{Wavelet‑based encoder (3 stages):}  
The encoder consists of three stages (\(n = 0, 1, 2\)). At each stage, the current feature \(X_n\) is fused with the corresponding Stream‑1 output \(x_n\) (downsampled to the same spatial size via \(\Phi_{\text{down}}\)). The result is then processed by a transformer block \(\Omega_{\text{enc}}\) that applies WQA for query enhancement, producing a coarser feature map:
\[
X_{n+1} = \Omega_{\text{enc}}\bigl( \langle X_n,\, \Phi_{\text{down}}(x_n) \rangle \bigr), \quad n = 0,1,2.
\]
This injection of convolutional detail improves localisation without sacrificing global context. The encoder outputs a bottleneck feature \(X_3\) at \(1/8\) of the original resolution.

\noindent\textbf{Decoder with cross-dimensional gating (3 stages):}
The decoder mirrors the encoder with three stages ((n = 2, 1, 0)). The bottleneck encoder feature \(X_3\) is first processed by a bottleneck transformer block to generate the initial decoder feature \(Z_3\). At each decoder stage, the previous coarse feature \(Z_{n+1}\) is upsampled via \(\Psi_{\text{up}}\) to match the spatial resolution of the corresponding encoder feature \(X_n\). The upsampled feature is then fused with the gated encoder skip feature \(\omega(X_n)\) (where \(\omega\) denotes CDGA) and the corresponding Stream-1 feature \(x_{n+3}\), resampled to the target resolution using \(\Phi_{\text{down}}\). The concatenated features are compressed using a \(1\times1\) convolution \(\Lambda\) (from (4c) to (c):
\[
\tilde{Z}_{n} =
\Lambda
\Bigl(
\bigl\langle
\Psi_{\text{up}}(Z_{n+1}),
\omega(X_n),
Z_{n+1},
\Phi_{\text{down}}(x_{n+3})
\bigr\rangle
\Bigr).
\]
The fused representation is subsequently refined using a decoder transformer block \(\Omega_{\text{dec}}\), which also employs WQA for query enhancement, consistent with the encoder stages.

\[ Z_n = \Omega_{\text{dec}}(\tilde{Z}_{n})
\]
\noindent\textbf{Disparity prediction:}  
Finally, the full‑resolution feature \(Z_1\) is projected by a \(1\times1\) convolution \(\psi_{out}\) to produce the refined disparity residual, which is added to \(d_{init}\) to obtain the final disparity \(d_{final}\).

This design couples all six Stream‑1 scales with the six transformer blocks, explicitly combining convolutional localisation with global attention. The asymmetric CDGA gating suppresses unreliable features during fusion, and the entire refinement module remains lightweight, adding minimal computation while substantially improving edge sharpness and robustness in challenging regions.

\subsection{Proposed EHR Loss}
\label{sec:ehr}
\vspace{-1mm}

We propose the \textit{EHR Loss} (Edge-guided and High-Error Region Loss), a composite loss designed to improve disparity estimation in challenging regions such as occlusions, boundaries, and textureless areas. Unlike uniform L1 or L2 losses, EHR Loss explicitly emphasizes boundaries and hard-to-predict regions.

\noindent \textbf{Edge-Aware Disparity Loss (\(\mathcal{L}_{\text{edge}}\)):}  
To enhance boundary accuracy, we weight disparity errors using Canny edge maps from the ground-truth and predicted disparities, \(E_{\text{gt}}\) and \(E_{\text{pred}}\), with thresholds \((t_{\min}, t_{\max}) = (0.1, 0.3)\):
\begin{equation}
\mathcal{L}_{\text{edge}} = \frac{1}{1 + \sum E_{\text{gt}}} \sum E_{\text{gt}} \cdot |D_{\text{gt}} - D_{\text{pred}}|
+ 0.1 \frac{1}{1 + \sum E_{\text{pred}}} \sum E_{\text{pred}} \cdot |D_{\text{gt}} - D_{\text{pred}}|.
\end{equation}
The predicted edge term is given a small weight (0.1) to act as a soft regulariser: it encourages the network to produce sharp edges that are consistent with its own predictions, while the low weight prevents reinforcing potential false positives early in training.

\noindent \textbf{High-Error Region Disparity Loss (\(\mathcal{L}_{\text{her}}\)):}  
To focus on difficult regions, we select the top \(k\%\) pixels with the largest absolute disparity errors:
\begin{equation}
\mathcal{L}_{\text{her}} = \frac{1}{|P_{\text{her}}|} \sum_{(i,j) \in P_{\text{her}}} |D_{\text{gt}}^{i,j} - D_{\text{pred}}^{i,j}|.
\end{equation}
The top-\(k\) operation creates a binary mask and is non-differentiable; during backpropagation, gradients flow only through the selected pixel errors, while the mask itself is detached (stop-gradient). We set \(k=7\%\) based on a sensitivity analysis (see Subsection~\ref{subsec:ablation}), which shows that this value gives the best trade‑off between accuracy and stability.

\begin{figure}[t!]
    \centering
    \includegraphics[width=0.9\linewidth]{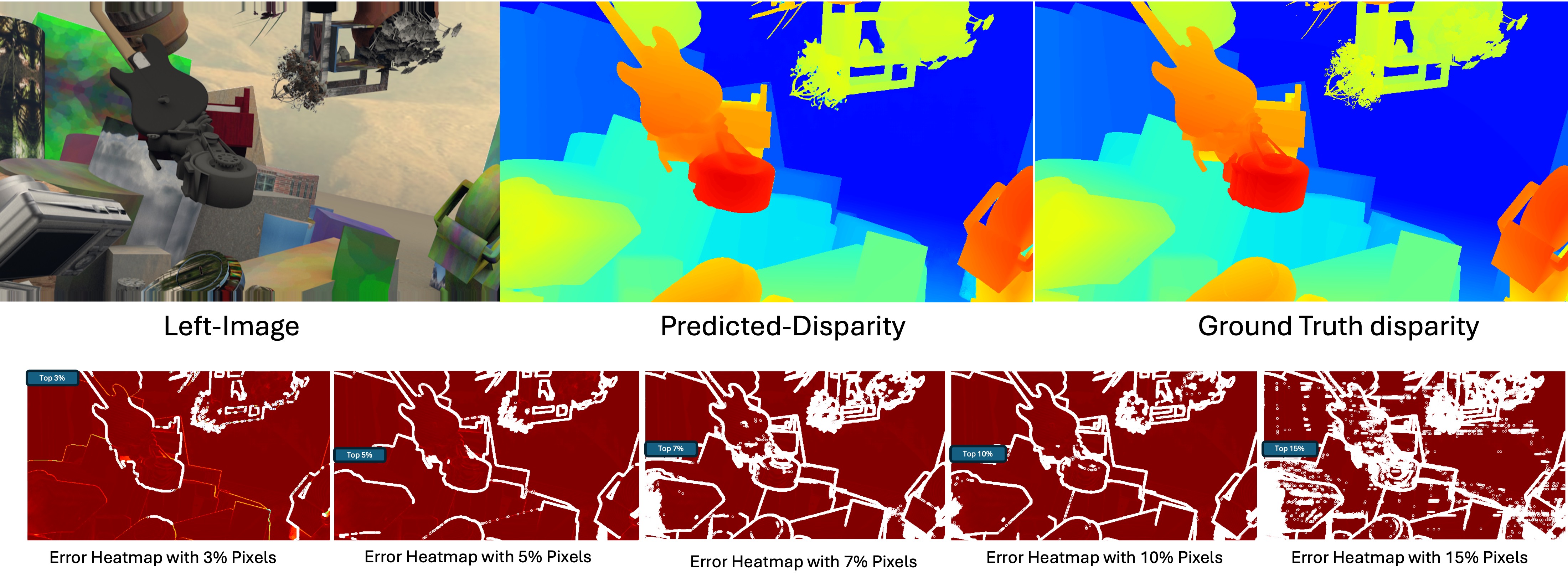}
    \caption{EHR Loss visualization. \textbf{Top:} Left input, predicted, and ground-truth disparity maps. \textbf{Bottom:} Error heatmaps for $k \in \{3, 5, 7, 10, 15\}\%$, showing $\mathcal{L}_{\text{her}}$'s error focus and $\mathcal{L}_{\text{edge}}$'s boundary enhancement.}
    \label{fig:ehr_visualization}
\end{figure}

\noindent \textbf{Combined Loss and Visualisation:}  
The final EHR loss is:
\begin{align}
\mathcal{L}_{\text{EHR}} = \lambda_1 \cdot \mathcal{L}_{\text{edge}} + \lambda_2 \cdot \mathcal{L}_{\text{her}},
\end{align}
where \(\lambda_1 = 1.0\) and \(\lambda_2 = 0.5\) balance boundary sharpening and hard-region focus. Figure~\ref{fig:ehr_visualization} visualises the effect of varying \(k\) and the edge-aware weighting, confirming that \(\mathcal{L}_{\text{her}}\) focuses on occluded and textureless regions while \(\mathcal{L}_{\text{edge}}\) sharpens boundaries. The total loss is given by the sum over all hierarchical refinement stages:
\begin{equation}
\mathcal{L}_{\text{total}} = \sum_{s=1}^{S} \left( \mathcal{L}_{1}^{s} + \mathcal{L}_{\text{EHR}}^{s} \right),
\end{equation}
where \(\mathcal{L}_{1}^{s}\) represents the L1 loss computed at scale \(s\) and \(S\) indicates the total number of refinement stages.

\begin{figure*}[!htb]
    \centering
    \includegraphics[width=0.95\linewidth]{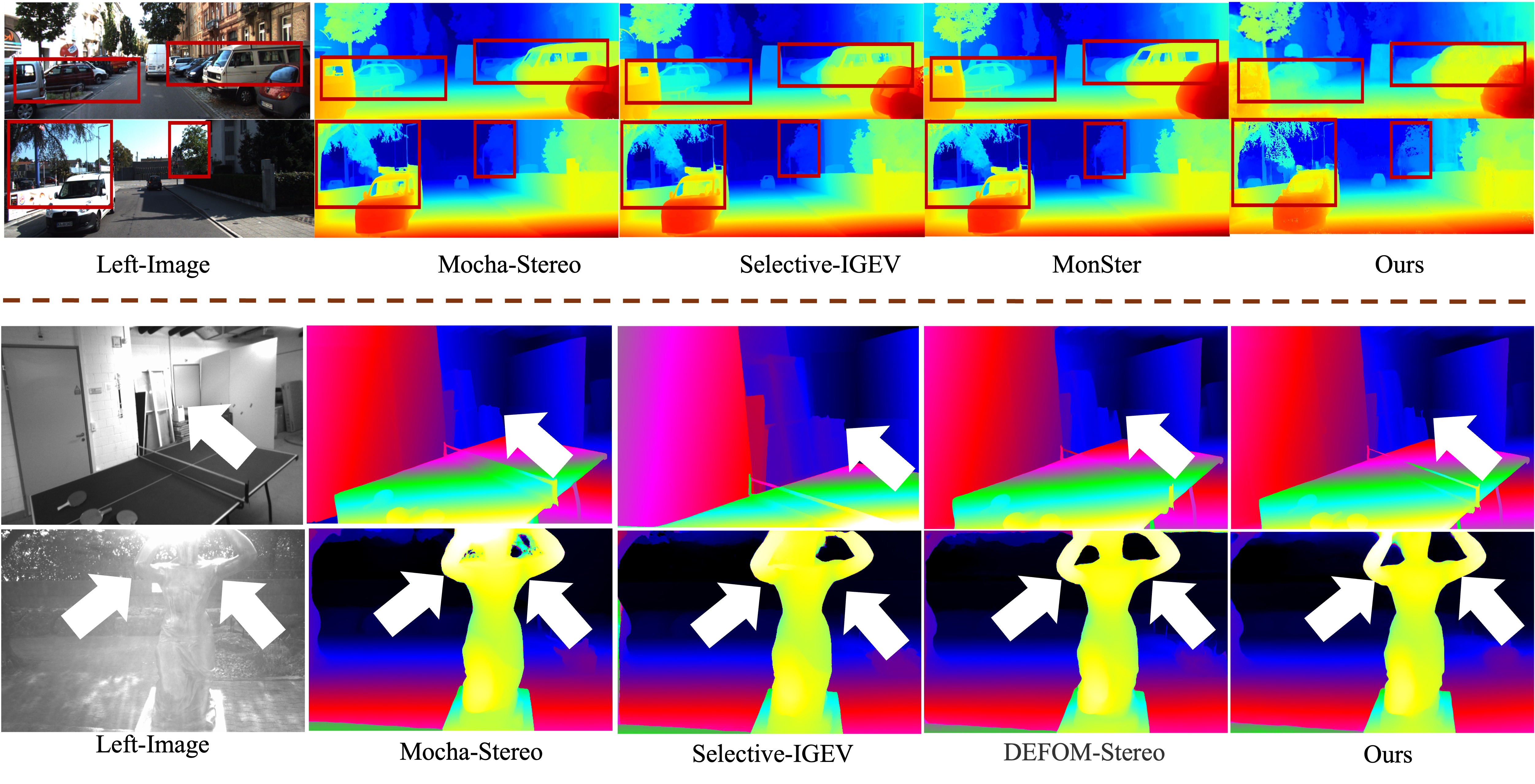}
    \caption{Visual comparison on the KITTI-15 (Zero-shot) on top and ETH3D (Fine-tuned) on bottom. From left to right: input left image, disparity predicted by SOTA methods, and disparity predicted by \protect\textit{PhasorNet}.}

    \label{fig:kitti_eth3d}
\end{figure*}

\noindent\textbf{Stage 2} freezes the feature and cost volume modules, and trains the refinement module with a multi-stage regression loss:
{\small
\begin{equation}
L_{\text{stage2}} = \mathit{Smooth}_{L1}(D_{\text{base}} - D_{\text{gt}}) + \sum_{j=1}^{n} \gamma^{n-j} \| D_j - D_{\text{gt}} \|_1,
\end{equation}
}
Here, $D_{\text{base}}$ denotes the base final output, $D_j$ are the intermediate predictions, $\gamma = 0.9$, $D_{\text{gt}}$ is the ground-truth disparity, and $n$ indicates the number of iterations. Each loss is used exclusively during its respective training phase. This hybrid supervision ensures stable, geometry-aware learning in Stage 1 and precise, high-resolution disparity in Stage 2, while maintaining real-time performance.

\subsection{Implementation Details}
\label{subsec:impl_details}
\vspace{-1mm}
\textit{PhasorNet} is implemented in PyTorch~\cite{paszke2019pytorch}. We train on the full Scene Flow dataset~\cite{mayer2016large} (35,454 stereo pairs) using the AdamW optimizer~\cite{loshchilov2017decoupled} with \(\beta_1 = 0.9\), \(\beta_2 = 0.999\), and weight decay \(1 \times 10^{-4}\). A OneCycle learning rate schedule~\cite{smith2017cyclical} with warmup and cosine annealing is applied, peaking at \(2 \times 10^{-4}\). Batch size is set to 4. Data augmentation and random cropping follow the Raft-Stereo protocol~\cite{lipson2021raft}. During evaluation, images are processed at their original resolution to comply with benchmark standards.
\begin{figure}[!htb]
    \centering
\includegraphics[width=1\linewidth]{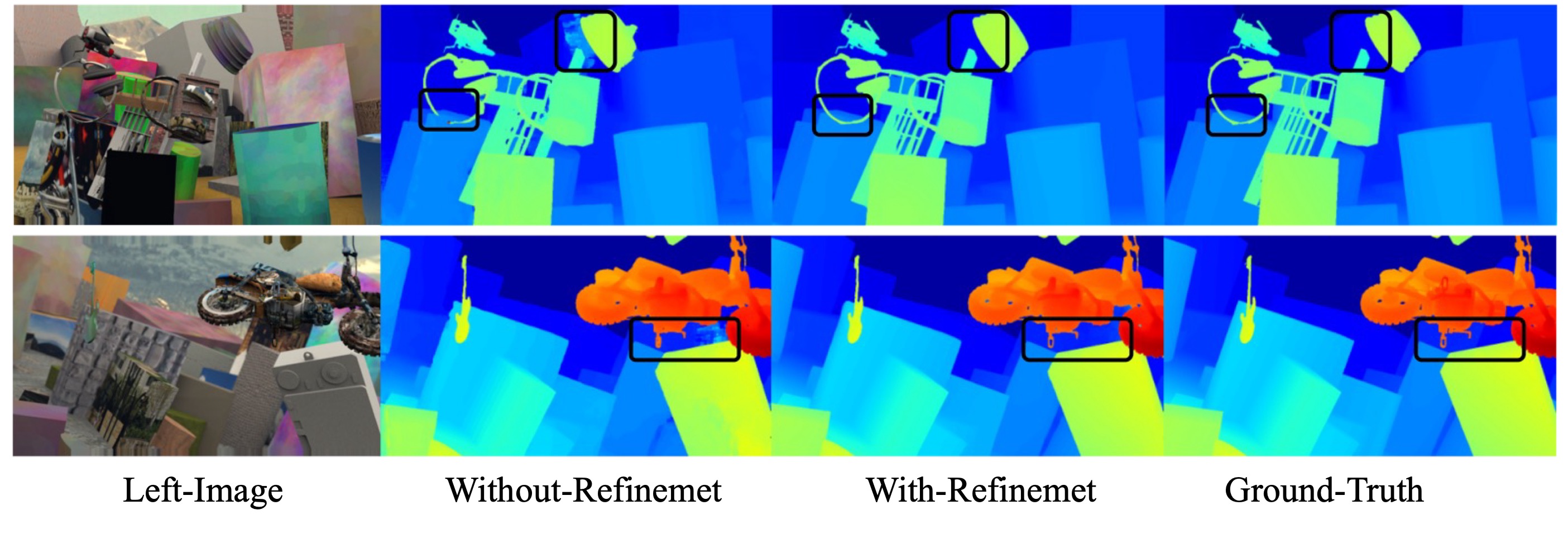}
    \caption{Visual results on the Scene Flow test set. From left to right: input left image, predicted disparity before and after refinement, and ground-truth disparity. The boxed region highlights improved boundary accuracy and detail due to refinement.}
    \label{fig:sf_visualization}
\end{figure}
\subsection{Datasets and Evaluation Metrics}
\vspace{-2mm}
\label{subsec:datasets_metrics}
We evaluate \textit{PhasorNet} on both synthetic and real-world datasets. \textbf{Scene Flow}~\cite{mayer2016large} provides 35,454 synthetic training pairs with dense disparity annotations and diverse textures, serving as the primary training set. For real-world evaluation, we use {KITTI-2012}~\cite{geiger2012we} (194 training, 195 testing) and {KITTI-2015}~\cite{menze2015object} (200 training, 200 testing), which include challenging driving scenes with occlusions, reflections, and sparse ground truth.{ETH3D}~\cite{schops2017multi} contains 27 training and 20 testing high-resolution stereo pairs from indoor and outdoor environments, featuring real-world difficulties such as textureless regions, reflective surfaces, thin structures, and varying illumination.

Performance is measured with standard metrics. On Scene Flow and KITTI, we report D1 Error (percentage of pixels with disparity errors $>$3 px or 5\% of ground truth). On ETH3D, we report Bad 1.0 (errors $>$1 px), AvgErr, and RMS, emphasizing large and boundary-related errors. Together, these metrics provide a comprehensive evaluation of \textit{PhasorNet}.

\begin{table*}[!htb]
\centering
\scriptsize
\setlength{\tabcolsep}{6pt}
\caption{Quantitative results on the ETH3D benchmark. \textbf{Bold} values indicate the best scores; \underline{underlined} values denote the second-best scores.}
\label{tab:eth3d}
\begin{tabular}{l|c|ccc|ccc}
\hline
\multirow{2}{*}{Method} & \multirow{2}{*}{Pub. Year} &
\multicolumn{3}{c|}{\textbf{ETH3D Non-occluded}} &
\multicolumn{3}{c}{\textbf{ETH3D All}} \\
 &  & Bad 1.0 & AvgErr & RMS & Bad 1.0 & AvgErr & RMS \\
\hline
RAFT-Stereo~\cite{lipson2021raft}              & 3DV-2021 & 2.44 & 0.18 & 0.36 & 2.60 & 0.19 & 0.42 \\
CREStereo~\cite{li2022practical}               & CVPR-2022 & 0.98 & 0.13 & 0.28 & {1.09} & 0.14 & 0.31 \\
CroCo-Stereo~\cite{weinzaepfel2023croco}       & CVPR-2023 & 0.99 & 0.14 & 0.30 & 1.14 & 0.15 & 0.35 \\
EAI-Stereo~\cite{zhao2022eai}                  & ACCV-2022 & 2.31 & 0.21 & 0.78 & 2.75 & 0.29 & 1.16 \\
IGEV-Stereo~\cite{xu2023iterative}             & CVPR-2023 & 1.12 & 0.14 & 0.34 & 1.51 & 0.20 & 0.86 \\
IGEV++~\cite{xu2025igev++}                     & TPAMI-2025 & 1.14 & 0.13 & 0.34 & 1.58 & 0.19 & 0.74 \\
Selective-IGEV~\cite{wang2024selective}        & CVPR-2024 & 1.23 & 0.12 & 0.29 & 1.56 & 0.15 & 0.57 \\
DEFOM-Stereo~\cite{jiang2025defom}             & CVPR-2025 & 0.70 & \underline{0.11} & 0.22 & 0.78 & \textbf{0.11} & \textbf{0.26} \\
MonSter~\cite{cheng2025monster}          & CVPR-2025 & \underline{0.44} & \textbf{0.10} & \textbf{0.20} & \underline{0.70} & \underline{0.13} & 0.47 \\
\rowcolor{gray!10} PhasorNet (Ours)             & --   & \textbf{0.41} & 0.12 & \underline{0.21} & \textbf{0.58} & \underline{0.13} & \underline{0.30} \\
\hline
\end{tabular}
\end{table*}

\begin{table*}[!htb]
\scriptsize
\setlength{\tabcolsep}{1pt}
\caption{Cross-dataset evaluation of models trained on Scene Flow (SF). We report D1 (Bad 3) error (\%, lower is better) on Scene Flow, KITTI-2012 (K-12), and KITTI-2015 (K-15). \textbf{Bold} and \underline{underlined} indicate the best and second-best results, respectively. Runtime (ms), parameter count (M), and memory usage (GB) are included for efficiency comparison.}

\label{tab:comparative1}
\centering
\setlength{\tabcolsep}{1pt}
\begin{tabular}{l c c c c c c}
\toprule
\textbf{Method} 
&\textbf{SF (D1)$\downarrow$}  &\textbf{K-12 (D1)$\downarrow$} 
& \textbf{K-15 (D1))$\downarrow$} 
& \textbf{Params (M)$\downarrow$} 
& \textbf{Time (ms)$\downarrow$} 
& \textbf{Mem (GB)$\downarrow$} \\
\midrule
RAFT-Stereo~\cite{lipson2021raft}          &6.08& 5.90 & 5.86 & \underline{11.11} & \underline{353} & 7.08 \\
IGEV-Stereo~\cite{xu2023iterative}        &5.21 & 5.19 & 6.06 & 12.60 & 410 & \underline{6.06} \\
DLNR~\cite{zhao2023high}                  & 5.06&  9.08 & 16.0 & 57.38 & 452 & 6.08 \\
Selective-IGEV~\cite{wang2024selective}    &4.98& 5.64 & 6.05 & 13.14 & 432 & 6.32 \\
Mocha-Stereo~\cite{chen2024mocha}          &2.35& {4.34} & 6.01 & 20.74 & 420 & 9.79 \\
MonSter~\cite{cheng2025monster}            &\underline{2.02}& \textbf{3.54} & \textbf{3.48} & 388.6 & 595 & 7.64 \\
\rowcolor{gray!10}
\textbf{Ours}                              &\textbf{1.95}& \underline{4.30} & \underline{4.19} & \textbf{5.300} & \textbf{108} & \textbf{2.08} \\
\bottomrule
\end{tabular}

\end{table*}

\begin{table}[t]
\centering
\scriptsize
\setlength{\tabcolsep}{1pt}
\caption{Zero-shot performance on challenging surfaces and edge preservation using KITTI-2012.  
\(D1_{\text{ref}}\) is the D1 error on the official reflective region mask (↓ lower is better).  
\(D1_{\text{edge}}\) is the D1 error restricted to pixels within 3px of ground‑truth disparity edges (↓ lower is better)}
\label{tab:targeted_metrics_horiz}
\begin{tabular}{l|ccccccc}
\hline
\rowcolor{gray!18} Metric & {RAFT-St.}\cite{lipson2021raft}  & {IGEV-St.}\cite{xu2023iterative}  & {DLNR}\cite{zhao2023high}  & {Sel.-IGEV}\cite{wang2024selective} & {Mocha-St.}\cite{chen2024mocha}  & {MonSter}\cite{cheng2025monster}  & \textbf{Ours} \\
\hline

\(D1_{\text{ref}}\) (\%) ↓ & 7.13  &  7.91
& 12.0 & 7.43 & 6.04 &\textbf{2.80} & \underline{4.95} \\
\(D1_{\text{edge}}\) (\%) ↓ & 6.32 & 7.33 & 10.8 & 7.68 & 6.40 & \underline{5.51} & \textbf{4.47} \\
\hline
\end{tabular}
\end{table}

\begin{table*}[!htb]
\centering
\scriptsize
\setlength{\tabcolsep}{1pt}
\caption{Single-pass zero-shot (trained on Scene Flow) performance comparison on KITTI-2012 and KITTI-2015. Bad3 error (\%, lower is better). All methods are evaluated in a constrained single-pass setting (1 iteration) to assess robustness under limited iterative refinement. Under this setting, MonSter degrades significantly (29.0\% on KITTI-2015), whereas PhasorNet remains accurate even in single-pass inference. \textbf{Bold} and \underline{underlined} indicate the best and second-best result}

\label{tab:comparative_single}
\centering
\setlength{\tabcolsep}{1pt}
\scriptsize
\begin{tabular}{l |c c c c c c c}
\hline
\rowcolor{gray!18}Metric Bad3 (\%) & {RAFT-St.}\cite{lipson2021raft}  & {IGEV-St.}\cite{xu2023iterative}  & {DLNR}\cite{zhao2023high}  & {Sel.-IGEV}\cite{wang2024selective} & 
{Mocha-St.}\cite{chen2024mocha}  & {MonSter}\cite{cheng2025monster}  & \textbf{Ours} \\
\hline
KITTI‑12  & 14.1 & \underline{6.06} & 30.5 & 6.78 & 6.95 & 20.1 & \textbf{5.16} \\
KITTI‑15 & 13.5 & 6.67 & 48.4 & \underline{6.68} & 6.80 & 29.0 & \textbf{5.72} \\
\hline
\end{tabular}
\end{table*}

\begin{table}[t]
\centering
\scriptsize
\caption{Comparison with lightweight stereo matching methods.
Lower is better for all error metrics.}
\label{tab:lightweight}
\begin{tabular}{l|c|cc|cc}
\hline
Method & Params (M) &
KITTI-12 & KITTI-15 &
ETH3D Non-occ. & ETH3D All \\
\hline
MobileStereoNet~\cite{shamsafar2022mobilestereonet} & 1.82 & 19.52 & 17.10 & -- & -- \\
HITNet~\cite{tankovich2021hitnet}         & 0.66 & 6.44  & 6.49  & 2.79 & 3.11 \\
BANet~\cite{xu2025banet}           & 3.63 & 17.92 & 17.61 & -- & -- \\
\rowcolor{gray!10}
PhasorNet (Ours)& 5.30 & \textbf{4.30} & \textbf{4.19}
                & \textbf{0.41} & \textbf{0.58} \\
\hline
\end{tabular}
\end{table}
\subsection{Results Analysis}
\vspace{-2mm}
We evaluate \textit{PhasorNet} on multiple standard benchmarks to assess accuracy, generalization, and efficiency. The results show a strong balance between high performance and low computational cost.

\noindent \textbf{ETH3D benchmark:} As shown in Table~\ref{tab:eth3d}, \textit{PhasorNet} achieves state-of-the-art performance on ETH3D with Bad 1.0 errors of \textbf{0.41\%} (non-occluded) and \textbf{0.58\%} (all pixels), surpassing MonSter (0.44\%, 0.70\%). It also ranks second in AvgErr (0.12) and RMS (0.21) on the non-occluded set, despite using only 5.3M parameters.

\noindent \textbf{Cross-Dataset Generalization and Targeted Evaluation:}
Table~\ref{tab:comparative1} reports cross-dataset performance. Trained on synthetic Scene Flow, \textit{PhasorNet} achieves a D1-all error of \textbf{1.95\%}, outperforming MonSter (2.02\%) and Mocha-Stereo (2.35\%). Without fine-tuning, it attains D1-all errors of 4.30\% on KITTI-2012 and 4.19\% on KITTI-2015. Furthermore, Table~\ref{tab:targeted_metrics_horiz} evaluates zero-shot performance on KITTI-2012 using reflective and edge masks. PhasorNet achieves ($D1_{ref}$ = 4.95\%) (second best) and ($D1_{edge}$ = 4.47\%) (best), outperforming all methods except MonSter on reflective regions. This confirms that our frequency-aware design improves robustness on reflective, transparent, and weakly textured surfaces while preserving fine boundaries.

\noindent \textbf{Single-Pass Inference.} Table~\ref{tab:comparative_single} presents single-pass (no iterative refinement) results. \textit{PhasorNet} achieves the lowest Bad 3.0 errors on KITTI-2012 (5.16\%) and KITTI-2015 (5.72\%), outperforming IGEV-Stereo and Selective-IGEV.

\noindent \textbf{Qualitative Results:} Figures~\ref{fig:kitti_eth3d} and~\ref{fig:sf_visualization} present qualitative comparisons on KITTI, ETH3D, and Scene Flow. \textit{PhasorNet} produces sharper and more coherent disparity maps, particularly along object boundaries and thin structures, while remaining robust in challenging regions such as reflective and transparent surfaces.

\subsection{Complexity Analysis}
\vspace{-2mm}

To assess computational efficiency, we benchmarked inference on a standard KITTI-2015 image pair (384 $\times$ 1248) using an NVIDIA RTX A6000 GPU. \textit{PhasorNet} achieves inference in just 108 ms per pair while requiring a mere 5.3M parameters and 2.08 GB  memory footprint. Compared to the heavy-weight MonSter architecture (388.6M parameters, 595 ms, 7.64 GB), our model delivers a 98\% fewer parameters and over 5$\times$ faster inference, demonstrating its strong viability for resource-constrained, real-world deployment.

To further substantiate our efficiency claims, Table~\ref{tab:efficiency_merged} provides a per-component latency analysis. 
The frequency-based modules (PAT, WQA, CDGA) together account for only \textbf{29.3\%} of the total inference time, 
while standard convolutional layers dominate the remaining {70.6\%}. 
FFT and DWT operations are parameter-free and memory-bound, contributing negligible computational cost compared to convolutions. 
Removing PAT or WQA reduces latency by only 10.7 ms (9.9\%) and 14.0 ms (12.9\%), respectively. 
The accuracy gains from adding these components (shown in Table~\ref{tab:ablation_combined}) are substantial: adding PAT, WQA, and CDGA progressively reduces refined EPE from 0.58 px to 0.52 px,
0.50 px, and finally 0.49 px. 
Moreover, our asymmetric WQA adds only 14 ms over the baseline, while a symmetric variant adds 45.6 ms and degrades accuracy (Table~\ref{tab:wqa_symmetric}), confirming the efficiency of our design.

\noindent \textbf{Iteration count:} 
All methods were evaluated using their default inference configurations (typically 32 iterations for recurrent approaches such as RAFT-Stereo, IGEV-Stereo, and MonSter). Despite using substantially fewer parameters and lower memory, PhasorNet achieves competitive or superior performance. To further evaluate robustness under limited refinement, we additionally assess all methods in a single-pass setting (1 iteration). Even under this constrained setup, PhasorNet achieves a Bad3 error of 5.72\% on KITTI-2015, whereas MonSter degrades significantly to 29.0\% (Table~\ref{tab:comparative_single}). These results suggest that PhasorNet maintains strong matching performance without relying heavily on many recurrent refinement iterations.) 
This demonstrates that PhasorNet’s accuracy does not rely on many iterative updates, making it far more suitable for real-time applications.

\noindent \textbf{Lightweight Baseline Comparison:} To complement the comparisons with recent state-of-the-art methods, we include MobileStereoNet, HITNet, and BANet as representative lightweight baselines. Although these methods use fewer parameters, PhasorNet achieves substantially lower KITTI-2012/2015 errors and better ETH3D accuracy than HITNet, highlighting its strong accuracy efficiency trade-off.

Overall, PhasorNet offers a favourable accuracy‑efficiency trade‑off, combining state-of-the-art methods ETH3D results with the lowest latency and parameter count among competing methods.

\begin{table}[t!]
\centering
\caption{Efficiency analysis on KITTI-2015 (384$\times$1248). Latency in ms.}
\label{tab:efficiency_merged}
\scriptsize
\setlength{\tabcolsep}{3pt}
\begin{tabular}{l|c|c|c|c}
\hline
\rowcolor{gray!18}\multicolumn{5}{c}{\textbf{Per-component breakdown}} \\
\hline
\textbf{Component / Variant} & PAT (FFT/IFFT) & WQA (DWT/IDWT) & CDGA & Backbone (conv) \\
\hline
Latency (ms) & 10.7 & 14.0 & 7.1 & 76.3  \\
Percentage (\%) & 9.9\% & 12.9\% & 6.5\% & 70.6\% \\
\hline
\rowcolor{gray!18}\multicolumn{5}{c}{\textbf{Speed ablation (removal impact)}} \\
\hline
Variant & PhasorNet & w/o PAT & w/o WQA & w/o CDGA  \\
\hline
Latency (ms) & 108.0 & 97.3 & 94.0 & 100.9  \\
\hline
\end{tabular}
\end{table}

\begin{table}[t]
\centering
\scriptsize
\setlength{\tabcolsep}{7pt}
\caption{Progressive ablation of PAC and GCFRM on Scene Flow. All variants use the full EHR Loss. Lower EPE is better.}
\label{tab:progressive_ablation}
\begin{tabular}{l|cc|c}
\hline
Variant & PAC & GCFRM & Final EPE $\downarrow$ \\
\hline
Baseline              & $\times$ & $\times$ & 0.58 \\
+ GCFRM               & $\times$ & \checkmark & 0.52 \\
\rowcolor{gray!10}
+ PAC (PhasorNet)      & \checkmark & \checkmark & \textbf{0.48} \\
\hline
\end{tabular}
\end{table}

\subsection{Ablation Study}
\label{subsec:ablation}
\vspace{-1mm}

To validate the effectiveness of PhasorNet's components, we conduct
ablation studies on the Scene Flow test set, reporting End-Point Error
(EPE) and D1 error (\% of pixels with error $>3$ px) for unrefined and
refined disparities.

Table~\ref{tab:progressive_ablation} presents the progressive ablation
of PAC and GCFRM, with all variants using the full EHR Loss. Adding
GCFRM reduces the final EPE from 0.58 to 0.52, while further
incorporating PAC improves it to 0.48. These progressive gains indicate
that PAC and GCFRM provide complementary rather than redundant
contributions.

Our full model (Ours in Table~\ref{tab:ablation_combined}) achieves a
refined D1 of 1.95\% and a refined EPE of 0.48. Removing phase
augmentation (w/o PAC) increases the refined D1 to 2.10\% and EPE to
0.52; removing wavelet query attention (w/o WQA) increases them to
2.05\% and 0.50, respectively; and removing cross-dimensional gated
attention (w/o CDGA) yields 2.00\% and 0.49. These consistent
degradations confirm the contribution of each component.

Table~\ref{tab:wqa_symmetric} shows that asymmetric WQA improves D1
over the no-WQA baseline (2.05$\rightarrow$1.98) and also outperforms
symmetric WQA (2.01) while requiring lower latency.
Table~\ref{tab:topk} further shows that $k=7\%$ yields the best refined
D1 of 1.95\%.

Finally, the loss ablation in Table~\ref{tab:ablation_combined} shows
that replacing $L_1$ supervision with the full EHR Loss reduces the
refined D1 from 1.98\% to 1.95\%, with the edge-aware and high-error
terms providing incremental gains. Overall, PhasorNet offers a favourable accuracy-efficiency trade-off,
combining state-of-the-art ETH3D performance with low latency and a compact parameter count compared with recent state-of-the-art methods.

\begin{table}[t!]
\centering
\scriptsize
\setlength{\tabcolsep}{6pt}
\caption{Architecture ablation (left): “w/o PAC” removes the FFT phase injection (standard cross‑attention); “w/o WQA” removes wavelet query attention; “w/o CDGA” removes cross‑dimensional gated attention. All architecture variants include the full EHR Loss. Loss ablation (right): component analysis of EHR Loss on the full PhasorNet architecture.}
\label{tab:ablation_combined}
\begin{tabular}{l|cc|cc||l|cc}
\toprule
\rowcolor{gray!18}\multicolumn{5}{c||}{\textbf{Architecture Component}} &
\multicolumn{3}{c}{\textbf{Loss Ablation}} \\
\midrule
\multirow{2}{*}{Variant} &
\multicolumn{2}{c|}{Unrefined} & \multicolumn{2}{c||}{Refined} &
Loss & EPE & D1 (\%) \\
& EPE & D1 (\%) & EPE & D1 (\%) &&& \\
\midrule
Ours (PhasorNet) & 0.58 & 2.30 & \textbf{0.48} & \textbf{1.95} &
L1 only       & 0.49   & 1.98 \\
w/o PAC       & 0.64 & 2.52 & 0.52 & 2.10 &
+ L\_edge (only)     & 0.49   & 1.97 \\
w/o WQA      & 0.62 & 2.45 & 0.50 & 2.05 &
+ L\_her (only)      & 0.485  & 1.96 \\
w/o CDGA              & 0.60 & 2.38 & 0.49 & 2.00 &
+ Full EHR      & \textbf{0.48} & \textbf{1.95} \\
\bottomrule
\end{tabular}
\end{table}

\begin{table}[!htb]
\centering
\scriptsize
\setlength{\tabcolsep}{1pt}
\begin{minipage}{0.48\textwidth}
\centering
\caption{Ablation of asymmetric (Q‑only) vs symmetric (Q,K,V) WQA on Scene Flow.}
\label{tab:wqa_symmetric}
\begin{tabular}{l|c|c}
\hline
Configuration & Refined D1 (\%) & Latency (ms) \\
\hline
Baseline (no WQA) & 2.05 & 94.0 \\
+ WQA (asymmetric) & \textbf{1.98} & 108 (+14) \\
+ WQA (symmetric) & 2.01 & 139.6 (+45.6) \\
\hline
\end{tabular}
\end{minipage}
\hfill
\begin{minipage}{0.45\textwidth}
\centering
\scriptsize
\setlength{\tabcolsep}{2pt}
\caption{Ablation study on the sensitivity of refined D1 (\%) to the choice of top‑k parameter in the high‑error region loss \(\mathcal{L}_{\text{her}}\).}
\label{tab:topk}
\begin{tabular}{c|c|c|c|c|c}
\hline
\(k\) & 3\% & 5\% & 7\% & 10\% & 15\% \\
\hline
Refined D1 (\%) & 1.97 & 1.96 & \textbf{1.95} & 1.96 & 1.98 \\
\hline
\end{tabular}
\end{minipage}
\end{table}

\section{Conclusion}
\vspace{-1mm}
We have presented PhasorNet, a lightweight stereo matching framework that integrates Fourier phase information and wavelet-based attention to improve robustness on challenging surfaces (reflective, transparent, weakly textured) while preserving fine details and boundaries. The Phase-Augmented Transformer (PAT) injects global phase-derived
structural cues that are robust to photometric variations into
cross-attention, while the Geometry-Context Fusion Refinement Module
(GCFRM) combines a convolutional stream with asymmetric wavelet
attention (WQA) and cross-dimensional gated attention (CDGA) for
efficient, detail-preserving refinement. A multi-scale EHR loss further focuses learning on edges and high-error regions. With only 5.3M parameters and 108 ms inference time on KITTI-2015, PhasorNet achieves state-of-the-art results on the ETH3D benchmark (Bad 1.0: 0.41\% non-occluded, 0.58\% all) and strong zero-shot generalisation on KITTI. Extensive ablations validate each component’s contribution. While PhasorNet improves robustness on weakly textured and reflective surfaces, perfectly uniform areas remain ambiguous for all correspondence-based methods, including ours. Future work will address such ambiguities without sacrificing efficiency.

\noindent\textbf{Acknowledgement:}\\
This research was supported by 
(d-real) Science Foundation Ireland (Grant 18/CRT/6224).
The authors thank the CVPR Lab members at Trinity Col-
lege Dublin and IIT Ropar for their support.

\bibliographystyle{splncs04}
\bibliography{main}

@String(CVPR= {IEEE Conf. Comput. Vis. Pattern Recog.})

@String(ICCV= {Int. Conf. Comput. Vis.})

@String(ICPR = {Int. Conf. Pattern Recog.})

@String(CVPR  = {CVPR})

@String(ICCV  = {ICCV})

@String(ICPR  = {ICPR})

@inproceedings{paszke2019pytorch,
  title={PyTorch: An Imperative Style, High-Performance Deep Learning Library},
  author={Paszke, Adam and Gross, Sam and Massa, Francisco and Lerer, Adam and Bradbury, James and Chanan, Gregory and Killeen, Trevor and Lin, Zeming and Gimelshein, Natalia and Antiga, Luca and others},
  booktitle={Advances in Neural Information Processing Systems (NeurIPS)},
  year={2019}
}

@article{dosovitskiy2020image,
  title={An image is worth 16x16 words: Transformers for image recognition at scale},
  author={Dosovitskiy, Alexey and Beyer, Lucas and Kolesnikov, Alexander and Weissenborn, Dirk and Zhai, Xiaohua and Unterthiner, Thomas and Dehghani, Mostafa and Minderer, Matthias and Heigold, Georg and Gelly, Sylvain and others},
  journal={arXiv preprint arXiv:2010.11929},
  year={2020}
}

@inproceedings{wang2022antioversmooth,
title={Anti-Oversmoothing in Deep Vision Transformers via the Fourier Domain Analysis: From Theory to Practice},
author={Wang, Peihao and Zheng, Wenqing and Chen, Tianlong and Wang, Zhangyang},
booktitle={International Conference on Learning Representations},
year={2022},
url={https://openreview.net/forum?id=O476oWmiNNp},
}

@inproceedings{zhao2023high,
  title={High-frequency stereo matching network},
  author={Zhao, Haoliang and Zhou, Huizhou and Zhang, Yongjun and Chen, Jie and Yang, Yitong and Zhao, Yong},
  booktitle={Proceedings of the IEEE/CVF conference on computer vision and pattern recognition},
  pages={1327--1336},
  year={2023}
}

@inproceedings{mayer2016large,
  title={A large dataset to train convolutional networks for disparity, optical flow, and scene flow estimation},
  author={Mayer, Nikolaus and Ilg, Eddy and Hausser, Philip and Fischer, Philipp and Cremers, Daniel and Dosovitskiy, Alexey and Brox, Thomas},
  booktitle={Proceedings of the IEEE conference on computer vision and pattern recognition},
  pages={4040--4048},
  year={2016}
}

@inproceedings{wang2024selective,
  title={Selective-stereo: Adaptive frequency information selection for stereo matching},
  author={Wang, Xianqi and Xu, Gangwei and Jia, Hao and Yang, Xin},
  booktitle={Proceedings of the IEEE/CVF Conference on Computer Vision and Pattern Recognition},
  pages={19701--19710},
  year={2024}
}

@inproceedings{chen2024mocha,
  title={Mocha-stereo: Motif channel attention network for stereo matching},
  author={Chen, Ziyang and Long, Wei and Yao, He and Zhang, Yongjun and Wang, Bingshu and Qin, Yongbin and Wu, Jia},
  booktitle={Proceedings of the IEEE/CVF Conference on Computer Vision and Pattern Recognition},
  pages={27768--27777},
  year={2024}
}

@inproceedings{geiger2012we,
  title={Are we ready for autonomous driving? the kitti vision benchmark suite},
  author={Geiger, Andreas and Lenz, Philip and Urtasun, Raquel},
  booktitle={2012 IEEE conference on computer vision and pattern recognition},
  pages={3354--3361},
  year={2012},
  organization={IEEE}
}

@inproceedings{menze2015object,
  title={Object scene flow for autonomous vehicles},
  author={Menze, Moritz and Geiger, Andreas},
  booktitle={Proceedings of the IEEE conference on computer vision and pattern recognition},
  pages={3061--3070},
  year={2015}
}

@inproceedings{li2022practical,
  title={Practical stereo matching via cascaded recurrent network with adaptive correlation},
  author={Li, Jiankun and Wang, Peisen and Xiong, Pengfei and Cai, Tao and Yan, Ziwei and Yang, Lei and Liu, Jiangyu and Fan, Haoqiang and Liu, Shuaicheng},
  booktitle={Proceedings of the IEEE/CVF conference on computer vision and pattern recognition},
  pages={16263--16272},
  year={2022}
}

@article{loshchilov2017decoupled,
  title={Decoupled weight decay regularization},
  author={Loshchilov, Ilya and Hutter, Frank},
  journal={arXiv preprint arXiv:1711.05101},
  year={2017}
}

@inproceedings{chang2018pyramid,
  title={Pyramid stereo matching network},
  author={Chang, Jia-Ren and Chen, Yong-Sheng},
  booktitle={Proceedings of the IEEE conference on computer vision and pattern recognition},
  pages={5410--5418},
  year={2018}
}

@inproceedings{shen2021cfnet,
  title={Cfnet: Cascade and fused cost volume for robust stereo matching},
  author={Shen, Zhelun and Dai, Yuchao and Rao, Zhibo},
  booktitle={Proceedings of the IEEE/CVF conference on computer vision and pattern recognition},
  pages={13906--13915},
  year={2021}
}

@inproceedings{xu2020aanet,
  title={Aanet: Adaptive aggregation network for efficient stereo matching},
  author={Xu, Haofei and Zhang, Juyong},
  booktitle={Proceedings of the IEEE/CVF conference on computer vision and pattern recognition},
  pages={1959--1968},
  year={2020}
}

@inproceedings{klaus2006segment,
  title={Segment-based stereo matching using belief propagation and a self-adapting dissimilarity measure},
  author={Klaus, Andreas and Sormann, Mario and Karner, Konrad},
  booktitle={18th International Conference on Pattern Recognition (ICPR'06)},
  volume={3},
  pages={15--18},
  year={2006},
  organization={IEEE}
}

@inproceedings{lipson2021raft,
  title={Raft-stereo: Multilevel recurrent field transforms for stereo matching},
  author={Lipson, Lahav and Teed, Zachary and Deng, Jia},
  booktitle={2021 International Conference on 3D Vision (3DV)},
  pages={218--227},
  year={2021},
  organization={IEEE}
}

@article{yang2024depth,
  title={Depth anything v2},
  author={Yang, Lihe and Kang, Bingyi and Huang, Zilong and Zhao, Zhen and Xu, Xiaogang and Feng, Jiashi and Zhao, Hengshuang},
  journal={Advances in Neural Information Processing Systems},
  volume={37},
  pages={21875--21911},
  year={2024}
}

@article{cheng2025monster,
  title={MonSter: Marry Monodepth to Stereo Unleashes Power},
  author={Cheng, Junda and Liu, Longliang and Xu, Gangwei and Wang, Xianqi and Zhang, Zhaoxing and Deng, Yong and Zang, Jinliang and Chen, Yurui and Cai, Zhipeng and Yang, Xin},
  journal={arXiv preprint arXiv:2501.08643},
  year={2025}
}

@article{felzenszwalb2006efficient,
  title={Efficient belief propagation for early vision},
  author={Felzenszwalb, Pedro F and Huttenlocher, Daniel P},
  journal={International journal of computer vision},
  volume={70},
  number={1},
  pages={41--54},
  year={2006},
  publisher={Springer}
}

@inproceedings{kendall2017end,
  title={End-to-end learning of geometry and context for deep stereo regression},
  author={Kendall, Alex and Martirosyan, Hayk and Dasgupta, Saumitro and Henry, Peter and Kennedy, Ryan and Bachrach, Abraham and Bry, Adam},
  booktitle={Proceedings of the IEEE international conference on computer vision},
  pages={66--75},
  year={2017}
}

@inproceedings{tankovich2021hitnet,
  title={Hitnet: Hierarchical iterative tile refinement network for real-time stereo matching},
  author={Tankovich, Vladimir and Hane, Christian and Zhang, Yinda and Kowdle, Adarsh and Fanello, Sean and Bouaziz, Sofien},
  booktitle={Proceedings of the IEEE/CVF conference on computer vision and pattern recognition},
  pages={14362--14372},
  year={2021}
}

@inproceedings{xu2023iterative,
  title={Iterative geometry encoding volume for stereo matching},
  author={Xu, Gangwei and Wang, Xianqi and Ding, Xiaohuan and Yang, Xin},
  booktitle={Proceedings of the IEEE/CVF conference on computer vision and pattern recognition},
  pages={21919--21928},
  year={2023}
}

@inproceedings{Zamir2021Restormer,
    title={Restormer: Efficient Transformer for High-Resolution Image Restoration}, 
    author={Syed Waqas Zamir and Aditya Arora and Salman Khan and Munawar Hayat 
            and Fahad Shahbaz Khan and Ming-Hsuan Yang},
    booktitle={CVPR},
    year={2022}
}

@inproceedings{tan2019efficientnet,
  title={Efficientnet: Rethinking model scaling for convolutional neural networks},
  author={Tan, Mingxing and Le, Quoc},
  booktitle={International conference on machine learning},
  pages={6105--6114},
  year={2019},
  organization={PMLR}
}

@inproceedings{schops2017multi,
  title={A multi-view stereo benchmark with high-resolution images and multi-camera videos},
  author={Schops, Thomas and Schonberger, Johannes L and Galliani, Silvano and Sattler, Torsten and Schindler, Konrad and Pollefeys, Marc and Geiger, Andreas},
  booktitle={Proceedings of the IEEE conference on computer vision and pattern recognition},
  pages={3260--3269},
  year={2017}
}

@article{xu2025igev++,
  title={Igev++: Iterative multi-range geometry encoding volumes for stereo matching},
  author={Xu, Gangwei and Wang, Xianqi and Zhang, Zhaoxing and Cheng, Junda and Liao, Chunyuan and Yang, Xin},
  journal={IEEE Transactions on Pattern Analysis and Machine Intelligence},
  year={2025},
  publisher={IEEE}
}

@inproceedings{he2016deep,
  title={Deep residual learning for image recognition},
  author={He, Kaiming and Zhang, Xiangyu and Ren, Shaoqing and Sun, Jian},
  booktitle={Proceedings of the IEEE conference on computer vision and pattern recognition},
  pages={770--778},
  year={2016}
}

@article{oppenheim1981importance,
  title={The importance of phase in signals},
  author={Oppenheim, Alan V and Lim, Jae S},
  journal={Proceedings of the IEEE},
  volume={69},
  number={5},
  pages={529--541},
  year={1981},
  publisher={IEEE}
}

@inproceedings{kuglin1975phase,
  author    = {Kuglin, Charles D. and Hines, D. C.},
  title     = {The Phase Correlation Image Alignment Method},
  booktitle = {Proceedings of the 1975 International Conference on Cybernetics and Society},
  pages     = {163--165},
  year      = {1975},
  organization = {IEEE}
}

@article{foroosh2002extension,
  author    = {Foroosh, Hassan and Zerubia, Josiane B. and Berthod, Marc},
  title     = {Extension of Phase Correlation to Subpixel Registration},
  journal   = {IEEE Transactions on Image Processing},
  volume    = {11},
  number    = {3},
  pages     = {188--200},
  year      = {2002},
  doi       = {10.1109/83.988953}
}

@inproceedings{xu2022attention,
  title={Accurate and Efficient Stereo Matching via Attention Concatenation Volume},
  author={Xu, Gangwei and Wang, Yun and Cheng, Junda and Tang, Jinhui and Yang, Xin},
  booktitle={Proceedings of the IEEE/CVF Conference on Computer Vision and Pattern Recognition},
  pages={12981--12990},
  year={2022}
}

@inproceedings{jiang2025defom,
  title={DEFOM-Stereo: Depth Foundation Model Based Stereo Matching},
  author={Jiang, Hualie and Lou, Zhiqiang and Ding, Laiyan and Xu, Rui and Tan, Minglang and Jiang, Wenjie and Huang, Rui},
  booktitle={IEEE International Conference on Computer Vision and Pattern Recognition (CVPR)},
  year={2025}
}

@article{li2022omni,
  title={Omni-dimensional dynamic convolution},
  author={Li, Chao and Zhou, Aojun and Yao, Anbang},
  journal={arXiv preprint arXiv:2209.07947},
  year={2022}
}

@article{smith2017cyclical,
  title={Cyclical learning rates for training neural networks},
  author={Smith, Leslie N},
  journal={2017 IEEE Winter Conference on Applications of Computer Vision (WACV)},
  pages={464--472},
  year={2017},
  organization={IEEE}
}

@inproceedings{weinzaepfel2023croco,
  title={Croco v2: Improved cross-view completion pre-training for stereo matching and optical flow},
  author={Weinzaepfel, Philippe and Lucas, Thomas and Leroy, Vincent and Cabon, Yohann and Arora, Vaibhav and Br{\'e}gier, Romain and Csurka, Gabriela and Antsfeld, Leonid and Chidlovskii, Boris and Revaud, J{\'e}r{\^o}me},
  booktitle={Proceedings of the IEEE/CVF International Conference on Computer Vision},
  pages={17969--17980},
  year={2023}
}

@inproceedings{zhao2022eai,
  title={Eai-stereo: Error aware iterative network for stereo matching},
  author={Zhao, Haoliang and Zhou, Huizhou and Zhang, Yongjun and Zhao, Yong and Yang, Yitong and Ouyang, Ting},
  booktitle={Proceedings of the Asian conference on computer vision},
  pages={315--332},
  year={2022}
}

@inproceedings{shamsafar2022mobilestereonet,
  title={Mobilestereonet: Towards lightweight deep networks for stereo matching},
  author={Shamsafar, Faranak and Woerz, Samuel and Rahim, Rafia and Zell, Andreas},
  booktitle={2022 IEEE/CVF Winter Conference on Applications of Computer Vision (WACV)},
  pages={677--686},
  year={2022},
  organization={IEEE}
}

@inproceedings{xu2025banet,
  title={Banet: Bilateral aggregation network for mobile stereo matching},
  author={Xu, Gangwei and Liu, Jiaxin and Wang, Xianqi and Cheng, Junda and Deng, Yong and Zang, Jinliang and Chen, Yurui and Yang, Xin},
  booktitle={2025 IEEE/CVF International Conference on Computer Vision (ICCV)},
  pages={28870--28880},
  year={2025},
  organization={IEEE}
}
\end{document}